\documentclass[journal]{IEEEtran}
\usepackage{graphicx}
\usepackage{cite}
\usepackage{amssymb}
\usepackage{amsmath}
\usepackage{enumerate}
\usepackage{booktabs}
\usepackage{algorithm}
\usepackage{algorithmic}
\usepackage{color}
\usepackage{colortbl}
\usepackage[colorlinks,linkcolor=red,anchorcolor=blue,citecolor=green]{hyperref}
\usepackage{bbm}
\usepackage{multirow}

\begin{document}

\title{Consistency-Regularized Region-Growing Network for Semantic Segmentation of Urban Scenes \\with Point-Level Annotations}

\author{Yonghao~Xu,~\IEEEmembership{Member,~IEEE,}
        and~Pedram~Ghamisi,~\IEEEmembership{Senior Member,~IEEE}
\thanks{Y. Xu is with the Institute of Advanced Research in Artificial Intelligence (IARAI), 1030 Vienna, Austria (e-mail: yonghaoxu@ieee.org).}
\thanks{P. Ghamisi is with the Institute of Advanced Research in Artificial Intelligence (IARAI), 1030 Vienna, Austria, and also with Helmholtz-Zentrum Dresden-Rossendorf, Helmholtz Institute Freiberg for Resource Technology, Machine Learning Group, 09599 Freiberg, Germany (e-mail: pedram.ghamisi@iarai.ac.at; p.ghamisi@hzdr.de).}
}

\markboth{Journal of \LaTeX\ Class Files,~Vol.~xx, No.~xx, June~2022}%
{Shell \MakeLowercase{\textit{et al.}}: Bare Demo of IEEEtran.cls for IEEE Journals}

\maketitle

\begin{abstract}
Deep learning algorithms have obtained great success in semantic segmentation of very high-resolution (VHR) remote sensing images. Nevertheless, training these models generally requires a large amount of accurate pixel-wise annotations, which is very laborious and time-consuming to collect. To reduce the annotation burden, this paper proposes a consistency-regularized region-growing network (CRGNet) to achieve semantic segmentation of VHR remote sensing images with point-level annotations. The key idea of CRGNet is to iteratively select unlabeled pixels with high confidence to expand the annotated area from the original sparse points. However, since there may exist some errors and noises in the expanded annotations, directly learning from them may mislead the training of the network. To this end, we further propose the consistency regularization strategy, where a base classifier and an expanded classifier are employed. Specifically, the base classifier is supervised by the original sparse annotations, while the expanded classifier aims to learn from the expanded annotations generated by the base classifier with the region-growing mechanism. The consistency regularization is thereby achieved by minimizing the discrepancy between the predictions from both the base and the expanded classifiers. We find such a simple regularization strategy is yet very useful to control the quality of the region-growing mechanism. Extensive experiments on two benchmark datasets demonstrate that the proposed CRGNet significantly outperforms the existing state-of-the-art methods. Codes and pre-trained models are available online (https://github.com/YonghaoXu/CRGNet).

\end{abstract}

\begin{IEEEkeywords}
Semantic segmentation, very high-resolution (VHR) images, weakly supervised learning, sparse annotation, convolutional neural network (CNN), remote sensing.
\end{IEEEkeywords}

\IEEEpeerreviewmaketitle

\section{Introduction}

\IEEEPARstart{S}{emantic} segmentation of very high-resolution (VHR) images is one of the most important tasks in the remote sensing field, which aims to produce a land-cover map by assigning a semantic label for each pixel in the image \cite{li2010edge}. Such high-resolution land-cover maps are essential to many fields of urban study \cite{ghamisi2019multisource,ratajczak2019automatic}, ranging from traffic analysis to urban planning \cite{sarkar2005landcover,xu2021self}.

The early study of semantic segmentation for VHR images mainly focuses on spatial or textural feature extraction \cite{tokarczyk2014features}. Some representative work includes morphological profiles (MPs) \cite{pesaresi2001new}, gray-level co-occurrence matrix (GLCM) \cite{zhang2017study}, wavelet transform \cite{myint2001robust}, and Gabor filter \cite{reis2011identification}. Generally, the extracted features will then be sent to a classifier like the support vector machine (SVM) or random forest (RF) to achieve pixel-wise land-cover mapping \cite{tokarczyk2014features}.

\begin{figure}
  \centering
  \includegraphics[width=\linewidth]{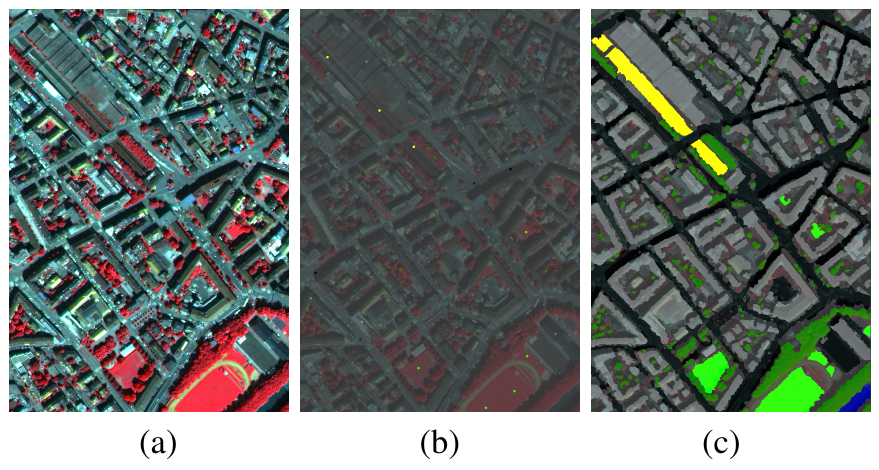}
  \caption{An illustration of different types of annotations for semantic segmentation of VHR images. The Zurich Summer dataset is used as an example. (a) The false color image. (b) Sparse point-level annotations used in this work. (c) Dense pixel-wise annotations used in previous study.}
\label{fig:pointsupervision}
\end{figure}

Witnessing the great success of deep learning algorithms in the computer vision field, recent research attempts to design advanced deep neural networks to tackle semantic segmentation of VHR satellite and aerial images \cite{maggiori2016convolutional,dfc,diakogiannis2020resunet}. Compared with hand-crafted features like MPs that depend largely on the prior information (empirical spatial filter parameters) of the designers, deep features can be automatically learned by the network without manual intervention, bringing about a better adaptation to different scenes \cite{ssun,hsic_review}. Nevertheless, since there are thousands of parameters that need to be learned in the deep neural networks, training these models usually requires a large amount of high-quality pixel-wise annotations, which is very laborious and time-consuming to collect in practice \cite{zhu_dlrs}. Once the training samples are insufficient, deep learning models may suffer from the over-fitting problem, resulting in a poor performance \cite{rpnet}.

\begin{figure*}
  \centering
  \includegraphics[width=\linewidth]{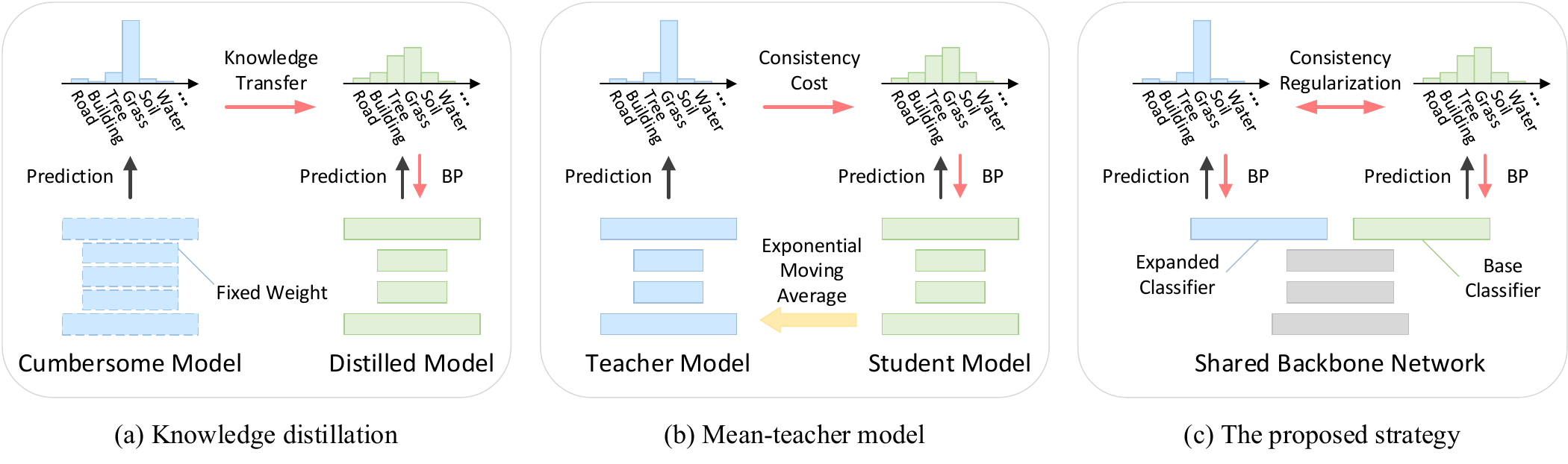}
  \caption{An illustration of different methods for knowledge transfer. (a) Knowledge distillation \cite{hinton2015distilling} adopts a cumbersome pre-trained model to conduct onesided knowledge transfer to the distilled model. (b) Mean-teacher model \cite{tarvainen2017mean} allows an onesided knowledge transfer from the teacher model to the student model with the consistency cost. (c) The proposed consistency-regularized strategy allows a bidirectional knowledge transfer for both the expanded classifier and the base classifier with the consistency regularization. ``BP'' denotes the back-propagation.}
\label{fig:demo}
\end{figure*}

The main burden of collecting accurate pixel-wise annotations for VHR remote sensing images lies in the boundary regions of different objects. As shown in \figurename~\ref{fig:pointsupervision} (c), due to the complex spatial distribution of the Earth's surface, annotating the detailed boundary for each object in the VHR image is very challenging, especially for those ambiguous regions \cite{hua2021semantic}. By contrast, the collection of point-level annotations is much easier for annotators since they only need to subconsciously mark some points inside the object without considering the detailed object boundary, as can be observed in \figurename~\ref{fig:pointsupervision} (b).

While point-level annotations could help to dramatically reduce the burden of collecting annotated data, directly training machine learning models with these highly sparse annotations would lead to very poor performance, especially for deep learning models that naturally require abundant training samples \cite{zhang_dlrs}. Thus, how to fully exploit the valuable information contained in the sparsely labeled VHR images is of crucial importance to the segmentation performance.

The initial inspiration of our method comes from an observation that adjacent pixels in remote sensing images, in particular those that are of high spatial resolution, tend to belong to the same category considering the spatial continuity of ground objects \cite{adams1994seeded}. Thus, a natural idea to tackle the insufficiency of annotations is to iteratively generate pseudo labels by expanding the annotated regions from the original sparse points with some well-designed criteria. The expanded annotations can then be used for training the network. As the annotated regions grow, the network could get stronger supervision, which in turn, helps to produce more accurate expansions in the region growing. Obviously, the segmentation performance of the whole framework is determined by the quality of the pseudo labels generated in the region growing. However, in practical applications, directly learning from the expanded annotations may misguide the training of the network because of the potential errors in the region growing, leading to worse segmentation results. This phenomenon may be even more serious for the semantic segmentation of VHR remote sensing images considering the high complexity of the spatial distribution of different objects.

To address the aforementioned challenge, this paper proposes a consistency-regularized region-growing network (CRGNet) for semantic segmentation of VHR images with point-level annotations. Specifically, the proposed CRGNet consists of a base classifier and an expanded classifier. In the training phase, the base classifier is supervised by the original sparse annotations, while the expanded classifier aims to learn from the expanded annotations generated by the base classifier with the region-growing mechanism. To make a balance between the learning of the original sparse annotations and the expanded annotations, we further propose a consistency regularization by minimizing the discrepancy between the predictions from both the base and the expanded classifiers. Despite its simplicity, the proposed regularization strategy can encourage a bidirectional knowledge transfer for both classifiers and is able to control the quality of the region-growing mechanism. Compared to existing knowledge transfer methods like the knowledge distillation \cite{hinton2015distilling} and mean-teacher model \cite{tarvainen2017mean}, the proposed strategy is more flexible and does not rely on external models as illustrated in \figurename~\ref{fig:demo}.

The main contributions of this study are summarized as follows.

\begin{enumerate}
\item A novel region-growing framework, namely CRGNet, is proposed for semantic segmentation of VHR remote sensing images with point-level annotations. With well-designed criteria, CRGNet can iteratively choose unlabeled pixels with high confidence to expand the annotated regions from the original sparse points, which helps to alleviate the insufficiency of training samples.
\item Since the accuracy of the expanded annotations can hardly be guaranteed, directly learning from them may misguide the training of the framework. To this end, a consistency regularization strategy is proposed. Specifically, we employ two classifiers including a base classifier and an expanded classifier in CRGNet, which are supervised by the original sparse annotations and the expanded annotations, respectively. The consistency regularization is then achieved by minimizing the discrepancy between the predictions of both classifiers.
\item We further conduct self-training with pseudo labels generated by the base classifier and the expanded classifier to finetune the proposed CRGNet. Extensive experiments on two challenging benchmark datasets demonstrate that the proposed CRGNet can yield competitive performance compared with the existing state-of-the-art approaches.
\end{enumerate}

The rest of this paper is organized as follows. Section II introduces some related work of this study. Section III describes the proposed CRGNet in detail. Section IV presents the information about datasets used in this study and the experimental results. Conclusions and other discussions are summarized in Section V.

\section{Related Work}
\subsection{Semantic Segmentation}
Semantic segmentation is a fundamental task for the interpretation of remote sensing data, which aims to assign a semantic label for each pixel in a given image. Inspired by the work in \cite{fcn}, many deep models have been proposed to tackle semantic segmentation of remote sensing images with fully convolutional networks (FCNs) \cite{lin2017maritime,mou2019relation,ssfcn}. In \cite{maggiori2016fully}, Maggiori \textit{et al.} adopted the FCN model to classify remote sensing images for the first time. Chen \textit{et al.}  proposed a symmetrical FCN framework with shortcut blocks for high-resolution remote sensing image semantic segmentation \cite{chen2018symmetrical}. Peng \textit{et al.} further proposed a multi-modal FCN for high-resolution remote sensing image, which incorporated the digital surface models (DSMs) using a dual-path architecture \cite{peng2019densely}.

Although the aforementioned deep learning models have achieved great success in semantic segmentation of remote sensing images, training these models generally requires a large amount of accurate pixel-wise annotations. However, in practical applications, the collection of such high-quality annotated data is very laborious and time-consuming \cite{hua2021semantic}. Thus, developing algorithms that can yield satisfactory segmentation performance with weak supervision (e.g., sparse point-level annotations) is of great significance.

\begin{figure*}
  \centering
  \includegraphics[width=\linewidth]{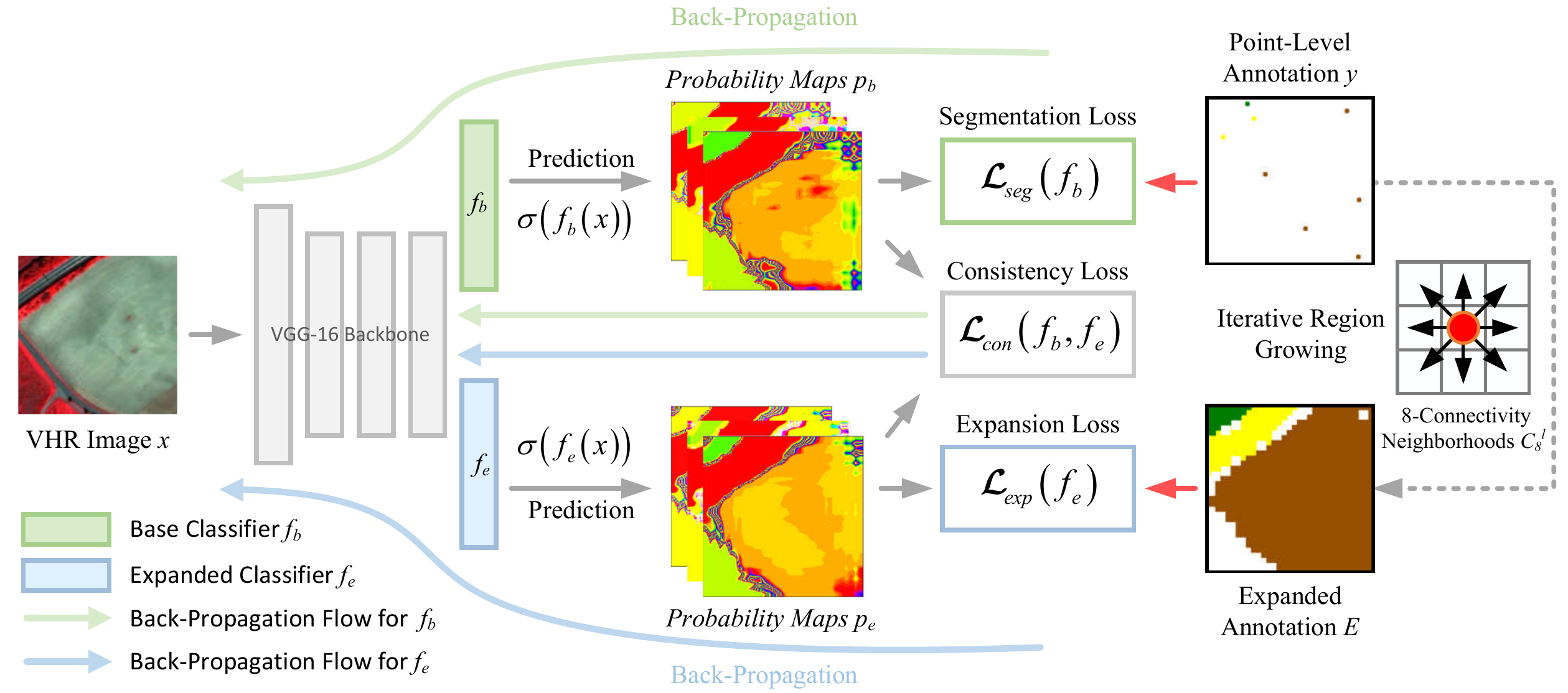}
  \caption{An illustration of the proposed consistency-regularized region-growing network (CRGNet) for weakly supervised semantic segmentation of VHR remote sensing images. There are two classifiers employed in the CRGNet, including a base classifier $f_b$ and an expanded classifier $f_e$. Both classifiers share the same backbone network. We use the original point-level annotations to train the base classifier $f_b$, while the expanded classifier $f_e$ is supervised by the expanded annotations generated with $f_b$ using the region-growing mechanism. We further let $f_b$ and $f_e$ learn from each other with a consistency regularization.}
\label{fig:CRGNet}
\end{figure*}

\subsection{Weakly Supervised Learning}
The success of most of the current state-of-the-art machine learning models depends largely on the sufficient ground-truth labels in the training, which are unattainable in many practical scenarios \cite{xu2019self}. To tackle this challenge, weakly supervised learning (WSL) methods are developed \cite{zhou2018brief}. Based on the type of supervision used in the model, WSL can be further divided into three categories. The first category is \textit{incomplete} supervision, where only a small subset of the training samples are labeled whereas the others are unlabeled \cite{hua2021semantic}. The second category is \textit{inexact} supervision, where the provided annotations are not exactly the ones that are expected for the task \cite{wei2016stc}. A typical example could be land-cover mapping using annotations with a lower spatial resolution than the observed remote sensing images \cite{wsl_rs}. The third category is \textit{inaccurate} supervision, where the provided annotations contain errors and noises. For example, learning with label noise \cite{kang2020robust,rs_label_noise}. In this study, we mainly address the first category of the WSL problem, where the provided annotations are point-level (i.e., only some sparse points are annotated with semantic labels in each image).

\subsection{Weakly Supervised Semantic Segmentation}
Compared to the fully supervised semantic segmentation where high-quality pixel-wise annotations are commonly required, the annotation burden for weakly supervised semantic segmentation could be greatly reduced. In \cite{wei2016stc}, Wei \textit{et al.} proposed a simple to complex framework for weakly supervised semantic segmentation using image-level annotations. With the bottom-up salient object detection techniques, they generated saliency maps of simple images which have a clean background without any pixel-wise annotations. These saliency maps are then regarded as pseudo labels to assist the training of a segmentation network. Kolesnikov \textit{et al.} proposed to generate weak localization cues by classification activation maps (CAMs) \cite{kolesnikov2016seed}. These weak localization cues were then used as pixel-level supervision to train the segmentation network. Huang \textit{et al.} further proposed the deep seeded region growing framework where the localization cues from CAMs were used to initialize the seeded region growing algorithm to generate new pixel-level labels \cite{huang2018weakly}.

Considering the difficulty of collecting dense pixel-level annotations for remote sensing data, weakly supervised semantic segmentation naturally fits the situation in Earth observation tasks. In \cite{yao2016semantic}, Yao \textit{et al.} proposed to transfer the deep features learned from the tile-level annotated data for semantic annotation of high-resolution satellite images. Hua \textit{et al.} proposed a feature and spatial relational regularization method for weakly supervised semantic segmentation of VHR images, where point-, line-, and polygon-level annotations are used as the weak supervision, respectively \cite{hua2021semantic}. Since convolutional neural networks (CNNs) trained with sparse annotations have the tendency to smooth the detailed object boundaries, Maggiolo \textit{et al.} further proposed a semi-supervised conditional random field (CRF) model to exploit the intermediate activation maps in CNNs and refine the segmentation performance \cite{maggiolo2021semisupervised}. In \cite{li2020accurate}, a weakly supervised cloud detection network is proposed, where only the block-level labels are used to indicate the presence or the absence of cloud in the remote sensing image block. Li \textit{et al.} further proposed a novel objective function with multiple weakly supervised constraints for cross-domain remote sensing image semantic segmentation \cite{li2021learning}.

In contrast to the existing methods \cite{maggiolo2021semisupervised,li2020accurate}, we propose to exploit the spatial continuity of ground objects that neighboring pixels tend to belong to the same category. By iteratively expanding the annotated regions from the original sparse points, our method could alleviate the problem of insufficiency of training samples. Besides, the proposed method enables the bidirectional knowledge transfer with the consistency regularization of two classifiers. Compared to the existing knowledge transfer methods \cite{hinton2015distilling,tarvainen2017mean}, the proposed strategy is a more moderate way to exploit the beneficial information in the expanded annotations.

\section{Methodology}
\subsection{Overview of the Proposed Model}
The key idea of the proposed consistency-regularized region-growing network (CRGNet) is to iteratively select unlabeled pixels with high confidence to expand the annotated area from the original sparse points. However, since the expanded annotations may differ from the true labels of the corresponding pixels, directly learning from them may mislead the training of the network. To this end, we further propose the consistency regularization strategy.

As shown in \figurename~\ref{fig:CRGNet}, there are two classifiers employed in the proposed CRGNet, including a base classifier and an expanded classifier. Both classifiers share the same backbone network. In the training phase, the base classifier is supervised by the original sparse annotations, while the expanded classifier aims to learn from the expanded annotations generated by the base classifier with the region-growing mechanism. The consistency regularization is then achieved by minimizing the discrepancy between the predictions from both classifiers. In the test phase, we average the predicted probability maps from both classifiers as the output of the whole framework.

\subsection{Region-Growing Mechanism}
One of the main challenges of weakly supervised semantic segmentation lies in the insufficiency of annotated samples. Considering the spatial continuity of ground objects that adjacent pixels are likely to belong to the same category, a natural idea is to expand the annotated area from the original sparse points with the region-growing mechanism.

Formally, let $f_b$ denote the mapping function of the base classifier. Given a VHR remote sensing image $x$, and the corresponding one-hot label $y$ (sparse point-level annotations), we first define the segmentation loss $\mathcal L_{seg}$ with the cross entropy for the base classifier $f_b$ as:
\begin{equation}
    \mathcal{L}_{seg}\left(f_b\right)=-\frac{1}{n_p}\sum_{i=1}^{n_p}\sum_{c=1}^{k}y^{\left(i,c\right)}\log\left(p_b^{\left(i,c\right)}\right),
\label{eq:ce}
\end{equation}
where $n_p$ and $k$ denote the number of annotated pixels in the original point-level annotation $y$, and the number of categories in the segmentation task, respectively. $p_b^{\left(i,c\right)}=\sigma\left(f_b\left(x\right)\right)^{\left(i,c\right)}$ denotes the probability of the $c$th class at pixel $i$ predicted by the base classifier $f_b$, and $\sigma\left(\cdot\right)$ denotes the softmax function. Note that we directly use $f_b$ as the input argument in \eqref{eq:ce} to represent that the optimization of $\mathcal{L}_{seg}$ is based on the parameters in $f_b$ for simplicity. Similar notations can be found in \eqref{eq:exp}, \eqref{eq:con}, and \eqref{eq:full}.

Recall that our goal is to expand the annotated regions. To this end, we define an expanded label matrix $E\in \left[0,k\right]$ ($1-k$ for $k$ categories and $0$ for the unlabeled pixels). At each iteration in the training phase, we first initialize $E$ with the original point-level label $y$:
\begin{equation}
    E^{\left(i\right)}=\mathop{\arg\max}\limits_{c}y^{\left(i,c\right)}.
\label{eq:initialization}
\end{equation}
Note that those unlabeled pixels in $E$ are simply set as $0$.

For each labeled pixel $l$ with $E^{\left(l\right)}>0$, let $C_8^{l}$ denote its corresponding $8$-connectivity neighborhood regions (we use $E^{\left(l\right)}$ to represent the value in the expanded label matrix at pixel $l$ for simplicity). Then, we visit each unlabeled pixel $u\in C_8^{l}$ with $E^{\left(u\right)}=0$, and update its label $E^{\left(u\right)}$ with the following criteria:
\begin{equation}
    E^{\left(u\right)} \leftarrow E^{\left(l\right)},~if~
    \begin{cases}
    \mathop{\arg\max}\limits_{c}\left(p_b^{\left(u,c\right)}\right)=E^{\left(l\right)}\\
    p_b^{\left(u,E^{\left(l\right)}\right)}\geq \tau,
    \end{cases}
\label{eq:RG}
\end{equation}
where we use ``$\leftarrow$'' to represent the right-to-left assigning operator, and $\tau$ is a probability confidence threshold parameter.

The first term of the criteria above ensures that the unlabeled pixel $u$ possesses the highest probability value in the same category ($E^{\left(l\right)}$) as the labeled pixel $l$. Since $u$ and $l$ are adjacent pixels, they likely belong to the same ground object in this case. The second term of the criteria further restricts that the probability value for the class $E^{\left(l\right)}$ should be greater than a confidence threshold $\tau$ considering that there may exist ambiguous categories which share very close probability values in the prediction of $p_b$. We repeat the updates in \eqref{eq:RG} until no pixel can satisfy the criteria.

\subsection{Consistency Regularization}
Once we obtain the expanded annotations $E$, a natural idea is to replace the original sparse label $y$ in \eqref{eq:ce} with the one-hot form of $E$ to train the segmentation network. Since there are more labeled samples in $E$, the network could get stronger supervision, which in turn, helps to produce more accurate expansions in the region growing. Nevertheless, directly learning from $E$ may misguide the training of the network due to the potential errors contained in $E$, leading to even worse segmentation results. Instead of directly training with $E$, in this subsection, we propose a novel consistency regularization strategy where the expanded classifier is employed to distill the supervised information contained in $E$.

Formally, let $f_e$ denote the mapping function of the expanded classifier. Note that both $f_b$ and $f_e$ share the same backbone network. Considering that objects with a larger spatial size tend to expand more pixels in the region growing, there may exist unbalance between different classes in the expanded annotations. Thus, we adopt the Lov$\rm{\acute{a}}$sz-Softmax loss \cite{berman2018lovasz} to train $f_e$ with $E$.

Specifically, let $\tilde{E}$ be the predicted label matrix of $f_e$:
\begin{equation}
    \tilde{E}^{\left(i\right)}=\mathop{\arg\max}\limits_{c}p_e^{\left(i,c\right)},
\label{eq:fe_e}
\end{equation}
where $p_e^{\left(i,c\right)}=\sigma\left(f_e\left(x\right)\right)^{\left(i,c\right)}$ denotes the probability of the $c$th class at pixel $i$ predicted by the expanded classifier $f_e$.

Then, the Jaccard index of class $c$ ($c\in \left[1,k\right]$) is defined as:
\begin{equation}
    J_c\left(\tilde{E},E\right)=\frac{|\{\tilde{E}=c\}\cap\{E=c\}|}{|\{\tilde{E}=c\}\cup\{E=c\}|}.
\label{eq:jc}
\end{equation}

The Jaccard index in \eqref{eq:jc} is also known as the intersection over union (IoU) metric. Since we expect the Jaccard index to increase in the training phase, the Jaccard loss $\Delta_{J_c}$ can thereby be defined as:
\begin{equation}
    \Delta_{J_c}\left(\tilde{E},E\right)=1-J_c\left(\tilde{E},E\right).
\label{eq:jc_loss}
\end{equation}

Considering that directly optimizing the Jaccard loss in \eqref{eq:jc_loss} is unfeasible, Berman \textit{et al.} proposed to approximate it with the prediction error tensor $M$ \cite{berman2018lovasz}, which can be defined as:
\begin{equation}
    M^{\left(i,c\right)}=
    \begin{cases}
    1-p_e^{\left(i,c\right)}~~~if~c=E^{\left(i\right)}\\
    p_e^{\left(i,c\right)}~~~~~~~~if~c\neq E^{\left(i\right)}.
    \end{cases}
\label{eq:error}
\end{equation}

\begin{algorithm}
    \caption{Training the proposed CRGNet}
    \label{alg:CRGNet}
 \begin{algorithmic}[1]
    \STATE Initialize the parameters in $f_b$ and $f_e$ with random Gaussian values.
    \FOR{$iter$ in $range\left(0,num\_iter\right)$}
    \STATE Get mini-batch samples $x$, $y$.
    \STATE Compute the probability map of $x$:\\ $p_{b}=\sigma\left(f_b\left(x\right)\right)$,  $p_{e}=\sigma\left(f_e\left(x\right)\right)$.
    \STATE Initialize the expanded label matrix $E$ via:\\ $E^{\left(i\right)}=\mathop{\arg\max}\limits_{c}y^{\left(i,c\right)}$.
    \STATE Initialize the flag variable $is\_grow\leftarrow True$.
    \WHILE{$is\_grow=True$}
    \STATE $\forall$ labeled pixel $l$, visit each unlabeled pixel $u\in C_8^{l}$.
    \STATE $is\_grow\leftarrow False$.
    \IF{$\mathop{\arg\max}\limits_{c}\left(p_b^{\left(u,c\right)}\right)=E^{\left(l\right)}$ and $p_b^{\left(u,E^{\left(l\right)}\right)}\geq \tau$}
    \STATE $E^{\left(u\right)} \leftarrow E^{\left(l\right)}$, and $is\_grow\leftarrow True$.
    \ENDIF
    \ENDWHILE
    \STATE Compute the segmentation loss $\mathcal L_{seg}\left(f_b\right)$, the expansion loss $\mathcal L_{exp}\left(f_e\right)$, and the consistency regularization loss $\mathcal L_{con}\left(f_b,f_e\right)$ via \eqref{eq:ce}, \eqref{eq:exp}, and \eqref{eq:con}.
    \STATE Compute the full loss function $\mathcal{L}\left(f_b,f_e\right)$ via \eqref{eq:full}.
    \STATE Update $f_b$ and $f_e$ by descending the stochastic gradients via     $\nabla_{f_b}\mathcal{L}\left(f_b,f_e\right)$ and $\nabla_{f_e}\mathcal{L}\left(f_b,f_e\right)$.
    \ENDFOR
    \STATE Compute the probability map $p$ for each training image $x$ via $p=\left(\sigma\left(f_b\left(x\right)\right)+\sigma\left(f_e\left(x\right)\right)\right)/2$, and finetune the network with the pseudo label matrix $E_p^{\left(i\right)}=\mathop{\arg\max}\limits_{c}p^{\left(i,c\right)}$.
\end{algorithmic}
\end{algorithm}

The expansion loss $\mathcal L_{exp}$ for the expanded classifier $f_e$ can thereby be formulated as:
\begin{equation}
    \mathcal{L}_{exp}\left(f_e\right)=-\frac{1}{n_e}\sum_{i=1}^{n_e}\sum_{c=1}^{k}\overline{\Delta_{J_c}}\left(M^{\left(i,c\right)}\right),
\label{eq:exp}
\end{equation}
where $n_e$ denotes the number of annotated pixels in the expanded annotation matrix $E$, and $\overline{\Delta_{J_c}}$ is the extended Jaccard loss. The detailed formulations for $\overline{\Delta_{J_c}}$ can be found in \cite{berman2018lovasz}. With the expansion loss in \eqref{eq:exp}, the expanded classifier $f_e$ can gradually get supervision from the expanded annotations.

Recall that our goal is to make a balance between the learning of the original sparse annotations and the expanded annotations. To this end, we further define a consistency regularization loss $\mathcal L_{con}$ by minimizing the discrepancy between the predictions from both the base and the expanded classifiers:
\begin{equation}
    \mathcal{L}_{con}\left(f_b,f_e\right)=-\frac{1}{n}\sum_{i=1}^{n}\sum_{c=1}^{k}\|p_b^{\left(i,c\right)}-p_e^{\left(i,c\right)}\|^2,
\label{eq:con}
\end{equation}
where $n$ denotes the number of pixels in the whole image.

Note that the consistency regularization loss $\mathcal L_{con}$ is applied to both the base and the expanded classifiers so that $f_b$ and $f_e$ can learn from each other.

The benefit of this loss function is twofold. First, although the expanded annotations can reduce the insufficiency of labeled samples, there may exist many errors and noises. By contrast, the original point-level annotations are accurate but highly sparse. Thus, the constraint in \eqref{eq:con} actually provides a balance between both annotations. Besides, $\mathcal L_{con}$ can be regarded as a soft knowledge distillation process \cite{hinton2015distilling}. With the help of the expanded classifier $f_e$, the base classifier $f_b$ no longer needs to directly learn from the expanded annotations. Instead, it is supervised by the distilled knowledge from $f_e$, which is a more moderate way to exploit the beneficial information in the expanded annotations.

\begin{figure*}
  \centering
  \includegraphics[width=\linewidth]{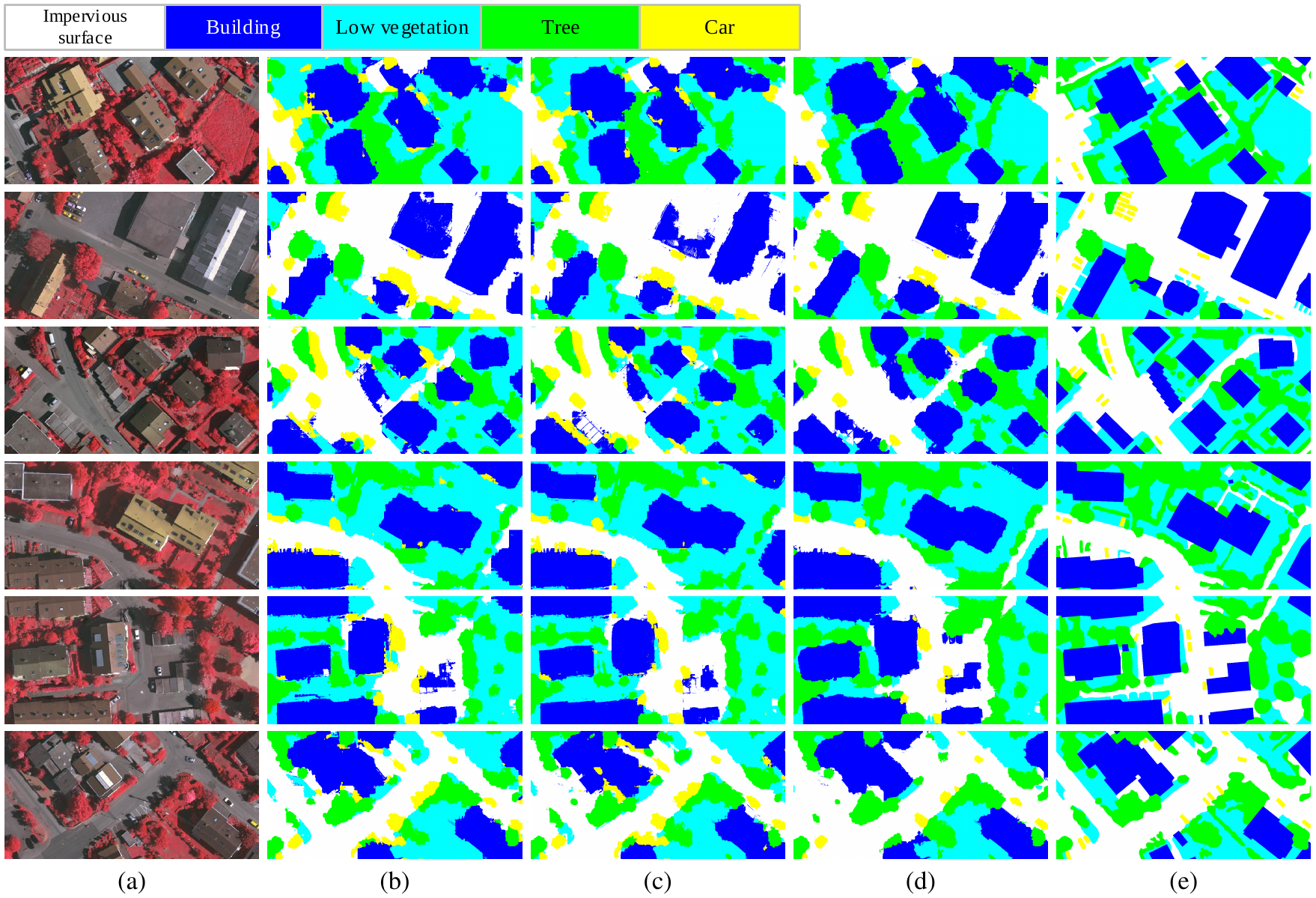}
  \caption{Qualitative semantic segmentation results for the Vaihingen dataset with different methods. (a) Input images from the Vaihingen dataset. (b) Baseline results with the vanilla DeepLab-v2 model. (c) Semi-supervised segmentation with the mean-teacher model. (d) The proposed CRGNet. (e) Ground-truth annotations.}
\label{fig:VaihingenMap}
\end{figure*}

The full loss function $\mathcal{L}$ for training the proposed CRGNet can be formulated as:
\begin{equation}
    \mathcal{L}\left(f_b,f_e\right)=\mathcal{L}_{seg}\left(f_b\right)+\mathcal{L}_{exp}\left(f_e\right)+\lambda_{con}\mathcal{L}_{con}\left(f_b,f_e\right),
\label{eq:full}
\end{equation}
where $\lambda_{con}$ is a weighting factor for the consistency regularization loss.

\subsection{Self-Training with Pseudo Labels}
Pseudo labeling is a commonly used technique in semi-supervised learning \cite{lee2013pseudo,song2020learning}. Different from previous self-training approaches which may require progressive selections for the most confident pseudo labels \cite{du2019ssf}, we simply conduct self-training with pseudo labels generated by $f_b$ and $f_e$ on the VHR images to finetune the proposed CRGNet, since the optimized $f_b$ and $f_e$ could already provide stable and high-quality pseudo labels.

Specifically, for each training image $x$, we obtain its probability map $p=\left(\sigma\left(f_b\left(x\right)\right)+\sigma\left(f_e\left(x\right)\right)\right)/2$.
The pseudo label matrix $E_p$ is generated with $E_p^{\left(i\right)}=\mathop{\arg\max}\limits_{c}p^{\left(i,c\right)}$. Then, we simply replace the expanded annotations $E$ with the pseudo label matrix $E_p$ to finetune the network by minimizing the objective function in \eqref{eq:full}.

The complete optimization procedure for the whole framework is shown in Algorithm~\ref{alg:CRGNet}. Note that for simplicity, the batch dimension is left out in the pseudo code.

\begin{figure*}
  \centering
  \includegraphics[width=\linewidth]{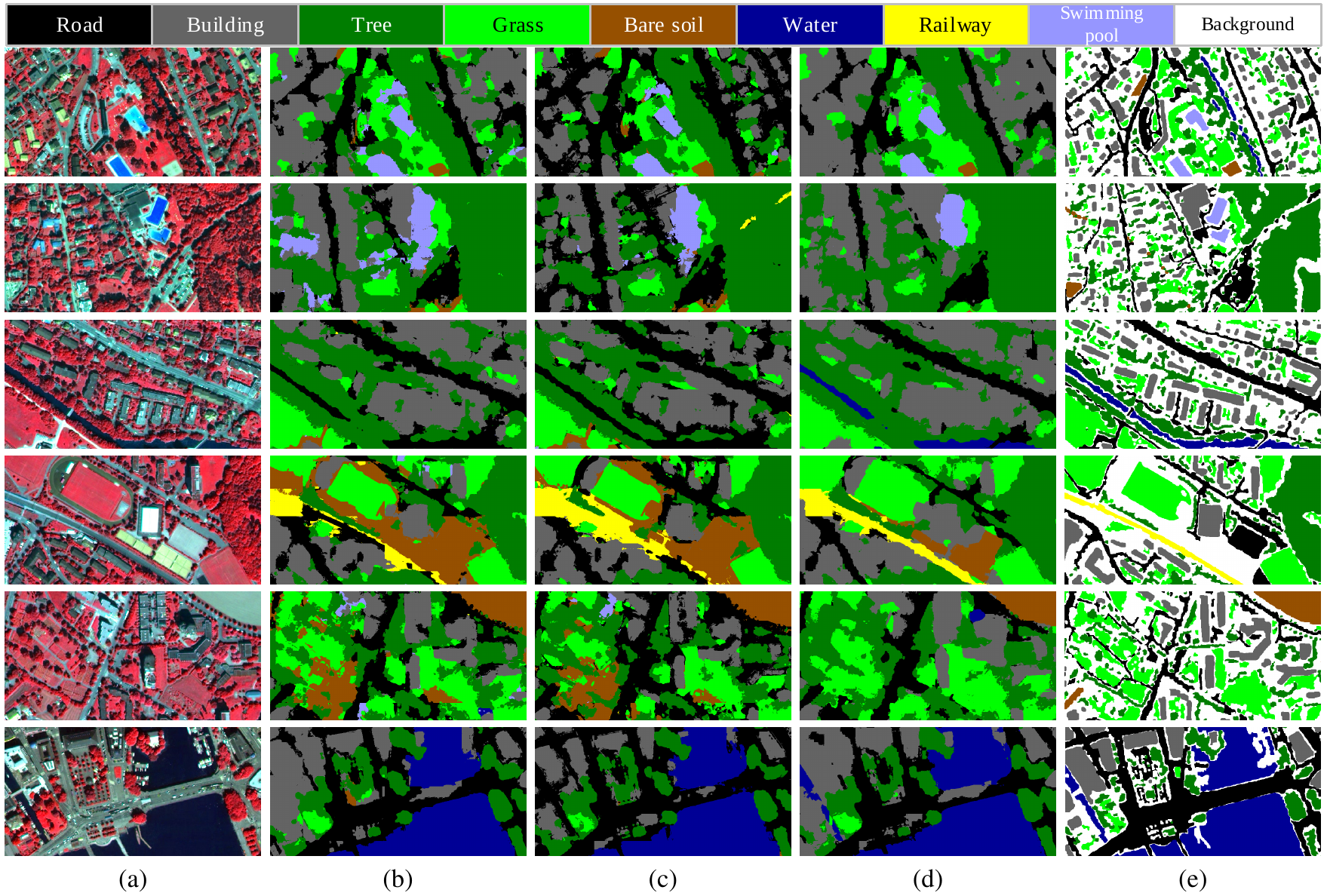}
  \caption{Qualitative semantic segmentation results for the Zurich Summer dataset with different methods. (a) Input images from the Zurich Summer dataset. (b) Baseline results with the vanilla DeepLab-v2 model. (c) Semi-supervised segmentation with the mean-teacher model. (d) The proposed CRGNet. (e) Ground-truth annotations.}
\label{fig:ZurichMap}
\end{figure*}

\section{Experiments}
\subsection{Data Descriptions}
Two benchmark VHR image datasets, including the Vaihingen\footnote{http://www2.isprs.org/commissions/comm3/wg4/2d-sem-label-vaihingen.
html} \cite{cramer2010dgpf} and the Zurich Summer \cite{volpi2015semantic} are utilized in this study.

\textbf{Vaihingen} is a benchmark dataset for semantic segmentation provided by the International Society for Photogrammetry and Remote Sensing (ISPRS), which is a subset of the data used for the test of digital aerial cameras carried out by the German Association of Photogrammetry and Remote Sensing (DGPF) \cite{cramer2010dgpf}. There are totally $33$ aerial images with a spatial resolution of $9$ cm collected over the city of Vaihingen. The average size of images is around $2500\times 1900$ pixels with a covering area of about $1.38$ km$^2$. For each aerial image, three bands are available, including the near-infrared, red, and green. Among these images, $16$ of them are fully annotated with $6$ different land-cover classes: impervious surface, building, low vegetation, tree, car, and clutter/background. To ensure experimental fairness, we
follow the same train-test split protocol as specified in the previous work \cite{hua2021semantic} and select five images (image IDs: 11, 15, 28, 30, 34) as the test set. The remaining ones are utilized to make up the training set.

\begin{table}
\centering
\caption{The numbers of the labeled pixels in point-level and pixel-wise annotations used in this study.}
\begin{tabular*}{\linewidth}{@{\extracolsep{\fill}}c|cc}
\toprule
Dataset&Point-level annotations&Pixel-wise annotations\\
\hline
Vaihingen$^*$&$18,787$&$54,373,518$\\
Zurich Summer&$29,508$&$12,266,287$\\
\bottomrule
\end{tabular*}
\\
\vspace{2pt}
\leftline{\quad $^*$Background/Clutter is not considered.}
\label{tab:number_pixel}
\end{table}

\textbf{Zurich Summer} consists of $20$ satellite images, which are taken over the city of Zurich in August 2002 by the Quick-Bird satellite \cite{volpi2015semantic}. The spatial resolution is $0.62$ m, and the average size of images is around $1000\times 1000$ pixels. The images consist of four channels, including the near-infrared, red, green, and blue. Following the previous work \cite{hua2021semantic}, we only utilize the near-infrared, red, and green channels in the experiments and select five images (image IDs: 16, 17, 18, 19, 20) as the test set. The remaining $15$ images are utilized to make up the training set. In total, there are $8$ urban classes, including road, building, tree, grass, bare soil, water, railway, and swimming pool. Uncategorized pixels are labeled as background.

In the training phase, we use the point-level annotations provided in \cite{hua2021semantic} as the supervision to train the proposed method. In the test phase, the full pixel-wise annotations from the original datasets are utilized for evaluation. The numbers of the labeled pixels in these two types of annotations are given in Table~\ref{tab:number_pixel}. It can be found that the point-level annotations are much fewer with several orders of magnitude than the original pixel-wise annotations.

\begin{table*}[!htb]
\centering
\caption{Quantitative results of semantic segmentation with point-level annotations on the Vaihingen dataset ($\%$). The per-class $F_1$ score, m$F_1$ score, mIoU, and OA are adopted as the performance metrics. Best results in each column are highlighted in \textbf{bold}.}
\resizebox{\textwidth}{!}{
\begin{tabular}{c|ccccc|ccc}
\toprule
Model & Impervious surface & Building & Low vegetation & Tree & Car & mF$_1$ & mIoU& OA\\
\hline
Baseline&68.30$\pm$3.06&78.14$\pm$1.68&61.64$\pm$0.82&75.20$\pm$0.72&27.36$\pm$7.09&61.63$\pm$1.84&46.96$\pm$1.57&68.97$\pm$1.74\\
Baseline+dCRF&73.45$\pm$2.86&78.15$\pm$3.31&60.73$\pm$2.19&71.97$\pm$6.37&39.78$\pm$5.89&64.82$\pm$1.57&50.49$\pm$0.89&72.46$\pm$1.23\\
MT&69.53$\pm$1.93&79.27$\pm$0.59&60.45$\pm$2.25&75.95$\pm$0.97&29.85$\pm$4.94&63.01$\pm$1.49&47.56$\pm$1.16&69.44$\pm$1.18\\
MT+dCRF&73.64$\pm$2.17&81.64$\pm$0.89&62.61$\pm$3.59&78.28$\pm$0.87&38.86$\pm$5.59&67.01$\pm$1.54&52.31$\pm$1.14&73.83$\pm$0.98\\
FESTA&74.65$\pm$2.73&78.64$\pm$4.74&60.24$\pm$3.33&76.15$\pm$2.07&23.65$\pm$4.24&62.66$\pm$2.54&--&71.43$\pm$2.93\\
FESTA+dCRF&\textbf{77.62$\pm$1.93}&80.08$\pm$5.27&60.78$\pm$4.00&76.70$\pm$2.00&31.40$\pm$5.24&65.32$\pm$2.56&--&73.65$\pm$2.52\\
CPS&70.83$\pm$1.66&78.77$\pm$2.15&60.82$\pm$1.06&76.46$\pm$0.95&33.02$\pm$5.42&63.98$\pm$1.83&49.09$\pm$1.71&70.75$\pm$1.40\\
CPS+dCRF&74.60$\pm$2.03&81.68$\pm$1.67&63.43$\pm$1.81&78.47$\pm$1.17&39.85$\pm$6.86&67.61$\pm$2.36&52.97$\pm$2.32&74.03$\pm$1.58\\
MixSeg&73.39$\pm$0.83&80.45$\pm$0.90&62.23$\pm$1.50&75.49$\pm$1.10&33.89$\pm$3.09&65.09$\pm$0.76&50.32$\pm$0.68&72.11$\pm$0.56\\
MixSeg+dCRF&77.51$\pm$0.96&82.03$\pm$1.46&64.08$\pm$2.91&77.75$\pm$1.00&45.46$\pm$3.36&68.87$\pm$0.40&54.64$\pm$1.12&75.29$\pm$0.87\\
CRGNet&73.88$\pm$1.29&81.43$\pm$0.77&65.36$\pm$0.57&77.84$\pm$0.91&41.86$\pm$4.46&68.07$\pm$1.39&52.95$\pm$1.83&74.11$\pm$1.16\\
CRGNet+dCRF&76.79$\pm$1.48&\textbf{82.46$\pm$1.07}&\textbf{66.59$\pm$1.20}&\textbf{79.73$\pm$0.67}&\textbf{49.04$\pm$4.90}&\textbf{70.92$\pm$1.36}&\textbf{55.96$\pm$1.81}&\textbf{76.24$\pm$0.98}\\
\hline
Oracle&86.21$\pm$0.09&91.26$\pm$0.07&74.96$\pm$0.16&84.61$\pm$0.06&72.68$\pm$0.39&81.94$\pm$0.07&70.01$\pm$0.09&84.37$\pm$0.08\\
\bottomrule
\end{tabular}
}
\label{tab:vaihingen}
\end{table*}

\begin{table*}[!htb]
\centering
\caption{Quantitative results of semantic segmentation with point-level annotations on the Zurich Summer dataset ($\%$). The per-class $F_1$ score, m$F_1$ score, mIoU, and OA are adopted as the performance metrics. Best results in each column are highlighted in \textbf{bold}.}
\resizebox{\textwidth}{!}{
\begin{tabular}{c|cccccccc|ccc}
\toprule
Model & Road & Build. & Tree & Grass & Soil & Water & Rail & Pool & mF$_1$& mIoU& OA\\
\hline
Baseline&67.06$\pm$3.31&75.34$\pm$2.32&79.20$\pm$1.29&72.54$\pm$2.65&39.35$\pm$7.63&87.29$\pm$0.31&14.45$\pm$12.71&45.79$\pm$12.01&60.13$\pm$2.29&47.73$\pm$2.44&72.45$\pm$3.04\\
Baseline+dCRF&73.43$\pm$3.57&80.98$\pm$2.54&84.54$\pm$1.80&79.02$\pm$2.47&53.01$\pm$10.21&91.37$\pm$0.39&12.17$\pm$18.20&62.94$\pm$14.01&67.18$\pm$2.62&55.06$\pm$2.16&75.85$\pm$3.21\\
MT&67.04$\pm$3.57&75.87$\pm$2.87&78.76$\pm$1.56&70.68$\pm$4.94&39.76$\pm$4.92&88.47$\pm$0.64&17.35$\pm$12.81&53.35$\pm$9.13&61.41$\pm$1.29&48.12$\pm$1.81&72.94$\pm$2.71\\
MT+dCRF&72.56$\pm$4.71&81.16$\pm$2.52&84.14$\pm$2.04&75.26$\pm$6.34&50.91$\pm$7.50&91.83$\pm$0.45&19.21$\pm$19.69&71.51$\pm$7.65&68.32$\pm$1.35&56.26$\pm$2.73&78.33$\pm$2.85\\
FESTA&70.64$\pm$3.44&77.34$\pm$4.13&82.91$\pm$2.48&83.73$\pm$2.34&56.67$\pm$5.64&89.67$\pm$2.25&0.94$\pm$1.89&73.62$\pm$4.06&66.94$\pm$2.56&--&78.17$\pm$3.00\\
FESTA+dCRF&71.23$\pm$2.61&77.71$\pm$3.17&82.81$\pm$1.99&\textbf{84.18$\pm$1.96}&66.34$\pm$3.69&93.40$\pm$1.81&0.00$\pm$0.00&77.38$\pm$8.87&69.05$\pm$1.15&--&79.11$\pm$2.14\\
CPS&67.62$\pm$5.65&76.76$\pm$4.56&79.37$\pm$3.08&73.15$\pm$6.09&52.04$\pm$7.30&87.09$\pm$0.77&9.33$\pm$10.75&67.35$\pm$5.93&64.09$\pm$2.53&50.79$\pm$3.11&74.56$\pm$3.57\\
CPS+dCRF&70.43$\pm$7.00&79.39$\pm$4.33&82.82$\pm$3.12&76.77$\pm$5.91&57.88$\pm$8.81&89.84$\pm$0.76&6.78$\pm$9.08&82.18$\pm$6.40&68.26$\pm$2.74&56.31$\pm$3.80&77.85$\pm$3.72\\
MixSeg&68.11$\pm$4.27&75.42$\pm$4.69&78.89$\pm$2.47&64.33$\pm$6.56&42.13$\pm$3.32&86.61$\pm$3.57&12.36$\pm$15.46&55.34$\pm$8.65&60.40$\pm$3.28&46.77$\pm$3.70&71.85$\pm$3.32\\
MixSeg+dCRF&73.75$\pm$4.69&80.23$\pm$3.31&83.66$\pm$2.08&68.36$\pm$9.92&46.48$\pm$11.98&90.44$\pm$4.32&21.18$\pm$21.10&77.37$\pm$7.48&67.68$\pm$2.83&56.03$\pm$2.97&77.36$\pm$3.31\\
CRGNet&70.77$\pm$2.14&79.36$\pm$1.43&80.90$\pm$0.98&79.16$\pm$1.46&59.96$\pm$1.20&90.95$\pm$1.41&28.57$\pm$14.53&80.45$\pm$3.96&71.26$\pm$2.37&58.08$\pm$3.52&76.89$\pm$2.75\\
CRGNet+dCRF&\textbf{75.42$\pm$2.35}&\textbf{81.86$\pm$0.64}&\textbf{85.75$\pm$0.83}&83.85$\pm$1.90&\textbf{69.57$\pm$4.30}&\textbf{93.62$\pm$1.30}&\textbf{29.39$\pm$23.93}&\textbf{86.01$\pm$4.11}&\textbf{75.68$\pm$2.54}&\textbf{63.40$\pm$3.91}&\textbf{80.36$\pm$3.29}\\
\hline
Oracle&88.78$\pm$0.09&93.47$\pm$0.09&92.08$\pm$0.75&87.86$\pm$1.36&64.53$\pm$2.36&94.94$\pm$0.37&12.51$\pm$7.52&84.29$\pm$3.41&77.31$\pm$0.90&68.63$\pm$1.02&90.00$\pm$0.41\\
\bottomrule
\end{tabular}
}
\label{tab:zurich}
\end{table*}

\subsection{Implementation Details}
We employ the DeepLab-v2 \cite{deeplab} with the VGG-16 model \cite{simonyan2014very} pre-trained on ImageNet \cite{imagenet} as the backbone networks. For the implementation of the base classifier $f_b$ and the expanded classifier $f_e$, we adopt the Atrous Spatial Pyramid Pooling (ASPP) \cite{deeplab} with dilation rates of $\{6,12,18,24\}$. The stochastic gradient descent (SGD) optimizer with a learning rate of $1e-3$ and a weight decay of $5e-5$ is utilized to train the model. We adopt the ``poly'' learning rate decay policy, where the initial learning rate is multiplied by $\left(1-iter/maxiter\right)^{power}$ with $power=0.9$ at each iteration. The number of total training iterations $maxiter$ is set to $5000$. After the pre-training phase, we further finetune the network with another $5000$ iterations using the self-training technique described in Section III-D. The $\tau$ in \eqref{eq:RG} and $\lambda_{con}$ in \eqref{eq:full} are empirically set to $0.95$ and $1$, respectively.

Due to the memory limit, we randomly crop the training images into $128\times 128$ patches, and the batch size in the training phase is set to $64$. In the test phase, we also crop the images into $128\times 128$ patches with a stride of $40$ pixels. Then, the segmentation maps of these patches are concatenated to achieve the complete land-cover mapping and evaluated with the ground-truth maps. We adopt the $F_1$ score per category, mean $F_1$ (m$F_1$) score, mean intersection over union (mIoU), and overall accuracy (OA) as the evaluation metrics.

The experiments in this paper are implemented in PyTorch with a single NVIDIA Tesla V100 GPU.
\subsection{Experimental Results}

\begin{figure*}
  \centering
  \includegraphics[width=\linewidth]{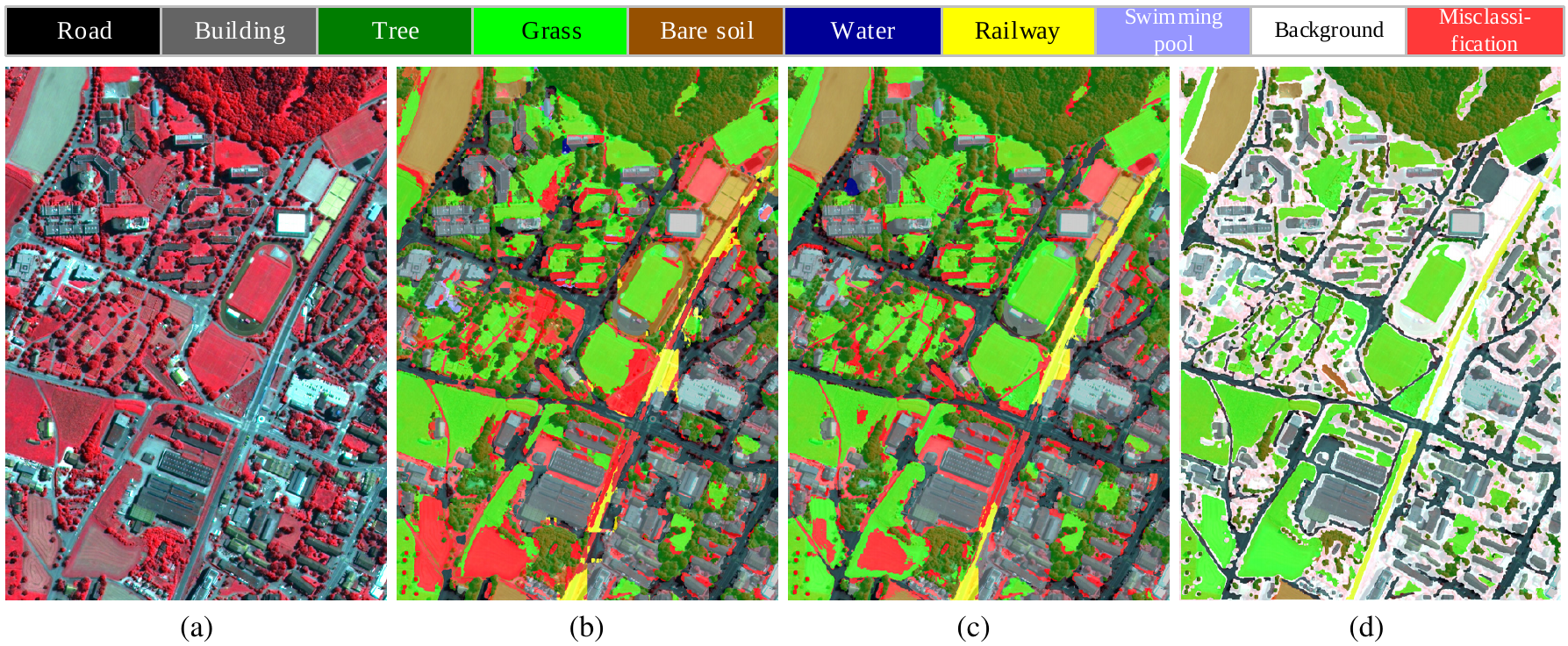}
  \caption{Example segmentation results of an image in the test set on the Zurich Summer dataset ($600,000~m^2$). (a) Input images from the Zurich Summer dataset. (b) Baseline results with the vanilla DeepLab-v2 model. (c) The proposed CRGNet. (d) Ground-truth annotations. The misclassification areas are denoted in \textcolor[rgb]{1.00,0.00,0.00}{red}. Zoom in for better visualization.}
\label{fig:ZurichFullMap}
\end{figure*}

\begin{figure*}
  \centering
  \includegraphics[width=\linewidth]{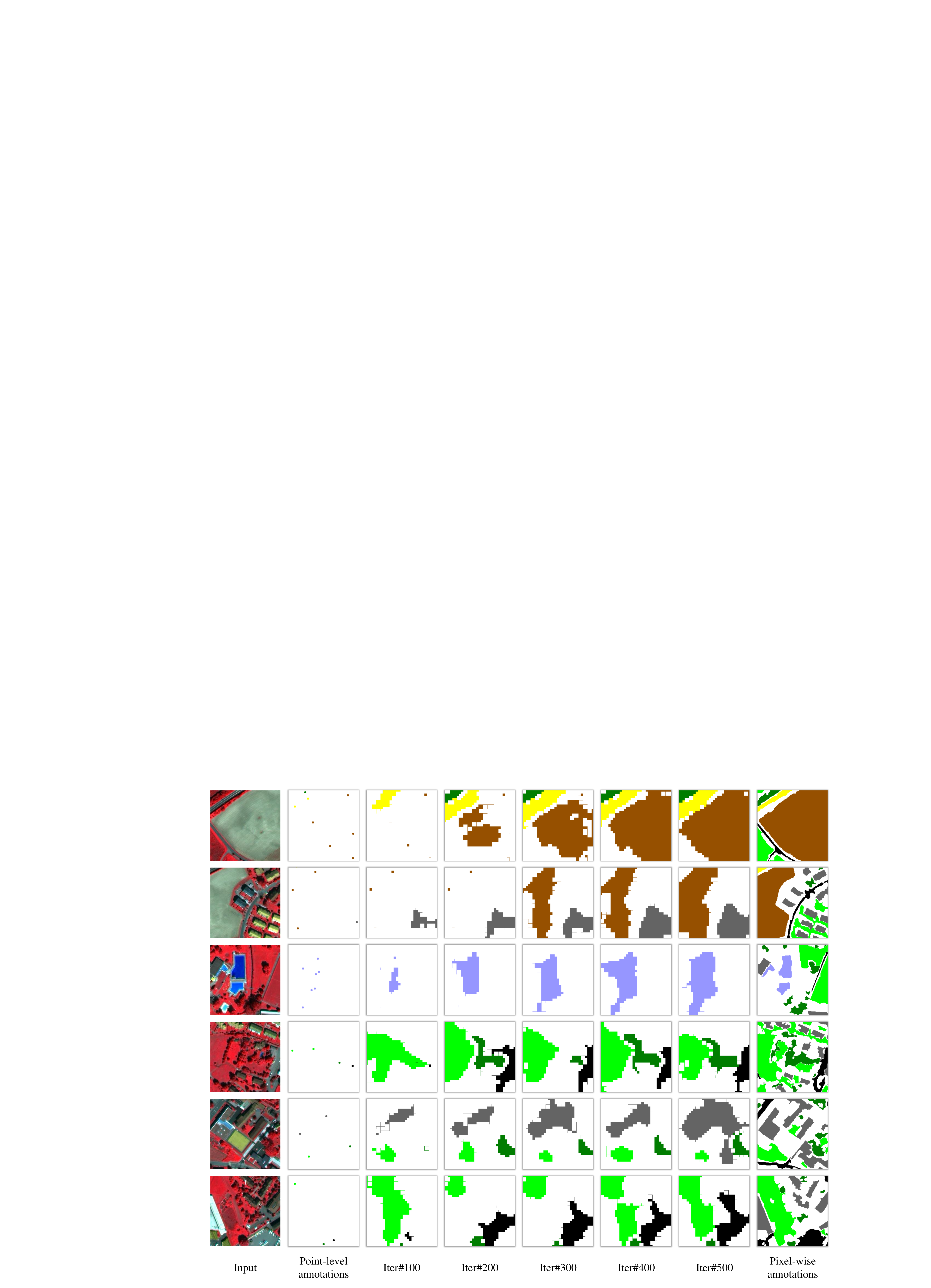}
  \caption{Dynamically expanded annotations in different iterations. Images from the Zurich Summer dataset are adopted as examples.}
\label{fig:exp}
\end{figure*}

In this subsection, we evaluate the performance of the proposed framework against recent state-of-the-art methods. All methods reported here adopt the VGG-16 model as the backbone network to ensure fair comparisons. A brief introduction to these methods is given below.

\begin{itemize}
\item Baseline: Segmentation with the vanilla DeepLab-v2 model \cite{deeplab}.
\item Baseline+dCRF: Segmentation with the vanilla DeepLab-v2 model and the post-processing of dense conditional random field (dCRF) \cite{maggiolo2018improving}.
\item MT: Semi-supervised segmentation with the mean-teacher (self-ensembling) model \cite{tarvainen2017mean}. The DeepLab-v2 model is adopted as the backbone network.
\item MT+dCRF: Segmentation with the mean-teacher model and the post-processing of dCRF.
\item FESTA: Segmentation with a novel feature and spatial relational regularization method \cite{hua2021semantic}.
\item FESTA+dCRF: Segmentation with the FESTA method and the post-processing of dCRF.
\item CPS: Semi-supervised segmentation with the cross pseudo supervision (CPS) using two segmentation networks perturbed with different initialization \cite{chen2021semi}.
\item CPS+dCRF: Segmentation with the CPS method and the post-processing of dCRF.
\item MixSeg: Semi-supervised segmentation with the CutMix augmentation techniques \cite{french2019semi}.
\item MixSeg+dCRF: Segmentation with the MixSeg method and the post-processing of dCRF.
\item CRGNet: Segmentation with the proposed consistency-regularized region-growing network.
\item CRGNet+dCRF: Segmentation with the proposed method and the post-processing of dCRF.
\item Oracle: Segmentation with the vanilla DeepLab-v2 model using full pixel-wise annotations.
\end{itemize}

\begin{table}
\caption{Performance contribution of each module in CRGNet (reported in mean F$_1$). Best results are highlighted in \textbf{bold}.}
\centering
\begin{tabular}{ccccc|cc} %
\toprule
Method&RG&CR&ST&dCRF&Vaihingen&Zurich Summer\\
\midrule
Baseline&&&&&61.63&60.13\\
$+$RG&$\checkmark$&&&&63.89&65.58\\
$+$CR&$\checkmark$&$\checkmark$&&&65.48&68.61\\
$+$ST&$\checkmark$&$\checkmark$&$\checkmark$&&68.07&71.26\\
$+$dCRF&$\checkmark$&$\checkmark$&$\checkmark$&$\checkmark$&\textbf{70.92}&\textbf{75.68}\\
\bottomrule
\end{tabular}
\label{tab:ablation}
\end{table}

The quantitative results are presented in Table~\ref{tab:vaihingen} and Table~\ref{tab:zurich}. Note that the results of the FESTA and FESTA+dCRF methods are directly duplicated from the original paper. It can be observed that the mean F$_1$ scores of the proposed CRGNet are about $7\%$ and $11\%$ higher than those of the baseline for the Vaihingen and Zurich Summer datasets, respectively (without the dCRF post-processing procedure). Besides, CRGNet can significantly outperform the recent state-of-the-art method FESTA around $5\%$ in the mean F$_1$ metric (without the dCRF post-processing procedure). We also find that the post-processing of dCRF plays an important role in improving the segmentation performance under the point-level supervision scenario. In both datasets, the performance gain obtained by the dCRF could reach around $2\%$ to $7\%$ for different methods. With the help of dCRF, the proposed CRGNet can rank first in $11$ categories and second in the left $2$ categories. These results verify the effectiveness of CRGNet in the weakly supervised semantic segmentation of VHR images.

Some qualitative results are presented in \figurename~\ref{fig:VaihingenMap} and \figurename~\ref{fig:ZurichMap}. We can find that the proposed CRGNet shows superiority in addressing those “hard examples” like the railway category in the Zurich Summer dataset. Due to the limited labeled railway samples contained in the training set, most methods yield relatively poor performance in this category (less than $20\%$ in the F$_1$ metric as shown in Table \ref{tab:zurich}). By contrast, the segmentation map of CRGNet is closer to the ground-truth annotations. It could achieve an F$_1$ score of about $28\%$ on “railway”, which outperforms the baseline with more than $14\%$.

To provide a thorough view of the performance of our method, we also exhibit a large-scale aerial scene as well as the corresponding semantic segmentation results in \figurename~\ref{fig:ZurichFullMap}. An image from the Zurich Summer dataset is adopted as an example. Those misclassification areas are colored in red. It can be observed that the misclassification areas are significantly reduced in the result of CRGNet, compared to the baseline method.

\subsection{Ablation Study}

To evaluate how each module in the proposed CRGNet influences the semantic segmentation performance, the quantitative ablation study results are demonstrated in Table \ref{tab:ablation}. Here, ``RG'' denotes the region growing mechanism, ``CR'' denotes the consistency regularization, ``ST'' denotes the self-training with pseudo labels, and ``dCRF'' denotes the dense CRF post-processing procedure. We use $+$RG to represent the method with constraints of both the segmentation loss $\mathcal{L}_{seg}\left(f_b\right)$ and the expansion loss $\mathcal{L}_{exp}\left(f_b\right)$ on the base classifier $f_b$ alone without using the expanded classifier $f_e$. In both datasets, directly applying RG only leads to limited mean F$_1$ gains, while combining both RG and CR can significantly improve the performance. ST also plays an important role in CRGNet, which improves the performance by more than $2\%$. Finally, with the help of dCRF, the mean F$_1$ scores got further increased, achieving state-of-the-art performance.

\begin{table}
\caption{Performance evaluation with different backbone networks (reported in mean F$_1$).}
\centering
\begin{tabular}{cccc}
\toprule
Method & Backbone Network & Vaihingen & Zurich Summer \\
\midrule
Baseline & \multirow{3}{*}{VGG-16} &61.63  &60.13  \\
CRGNet &  &68.07  &71.26  \\
Oracle &  &81.94  &77.31  \\
\midrule
Baseline & \multirow{3}{*}{ResNet-101} &61.95  &61.36  \\
CRGNet &  &69.18  &73.13  \\
Oracle &  &82.92  &78.29 \\
\bottomrule
\end{tabular}
\label{tab:backbone}
\end{table}

To analyze how different backbone networks would influence the performance of the proposed method, we have also conducted experiments using the ResNet-101 as the backbone network. It can be observed from Table \ref{tab:backbone} that using a stronger backbone network like the ResNet-101 can bring about limited improvement to the performance of the baseline method and the proposed CRGNet, especially on the Vaihingen dataset. By contrast, the oracle method can obtain an improvement of more than 2 percentage points on the mean F$_1$ metric for both datasets using the ResNet-101 as the backbone network. This phenomenon also indicates that the insufficiency of training samples is the main barrier to achieving satisfactory performance for the point-level semantic segmentation task.

To visually analyze the region-growing process in the proposed CRGNet, the dynamically expanded annotations in different iterations are visualized in \figurename~\ref{fig:exp}. Note that since we initialize the expanded label matrix $E$ in \eqref{eq:initialization} with the original point-level label $y$ at each iteration, there may exist inconsistency between the expanded annotations at different iterations. It can be observed that in the early iteration, there are relatively fewer samples selected in the region-growing mechanism because of the proposed confidence criterion. As the training goes on, the expanded annotations could gradually enlarge the labeled area from the original sparse points. For those scenes with simple spatial distributions (e.g., the first row in \figurename~\ref{fig:exp}), the proposed region-growing mechanism can eventually well simulate the dense ground-truth annotations. Nevertheless, there also exist some errors and noises in the expanded annotations in some complex scenarios. These results are also in accord with our intuition that directly learning from the expanded annotations may bring about inaccurate supervision to the framework.

\begin{figure}
  \centering
  \includegraphics[width=\linewidth]{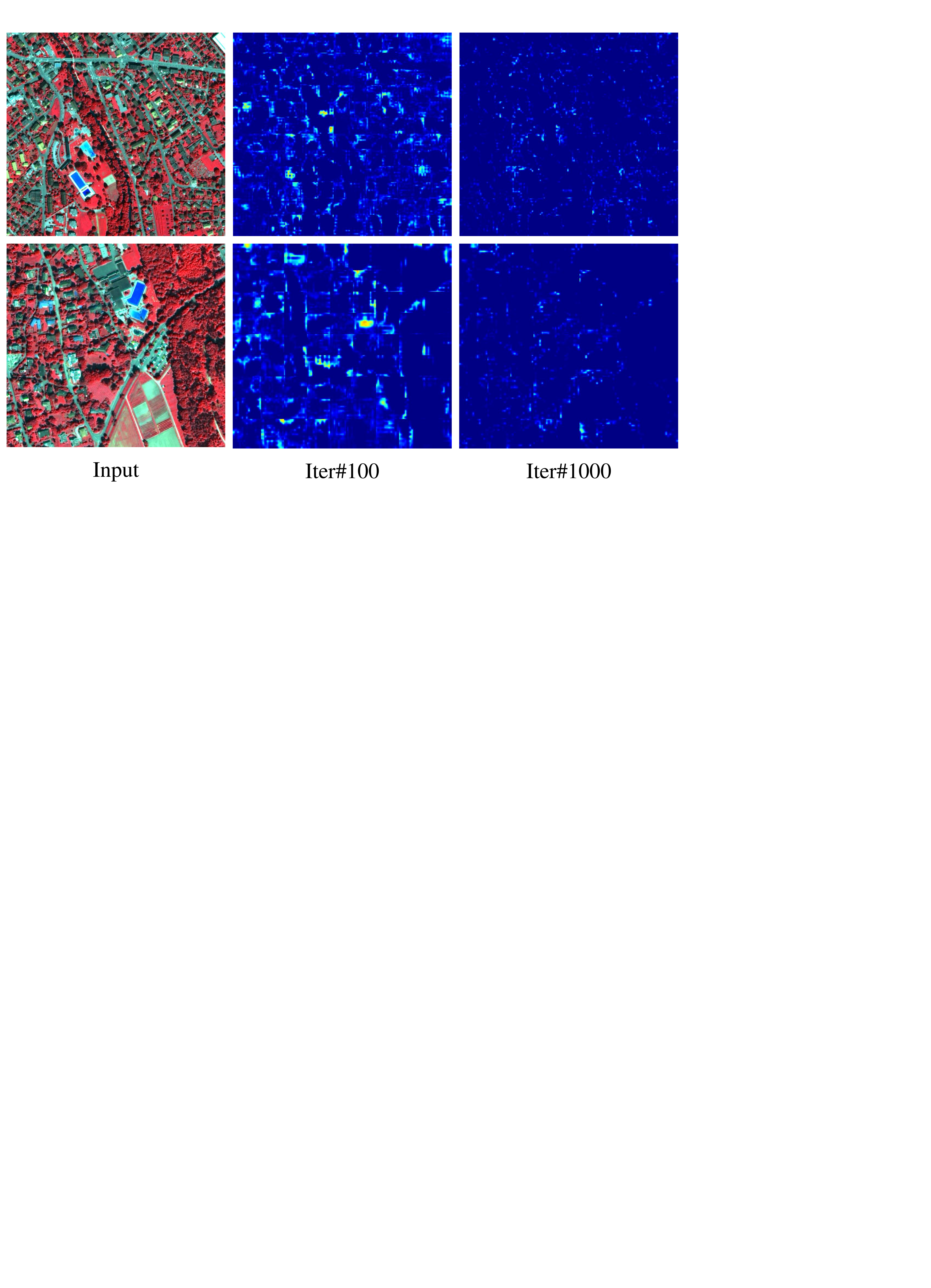}
  \caption{The prediction discrepancy between the base classifier $f_b$ and the expanded classifier $f_e$ in different iterations. Red regions in the map correspond to high discrepancy while blue ones correspond to low discrepancy. Images from the Zurich Summer dataset are adopted as examples.}
\label{fig:dif}
\end{figure}

\begin{figure}
  \centering
  \includegraphics[width=\linewidth]{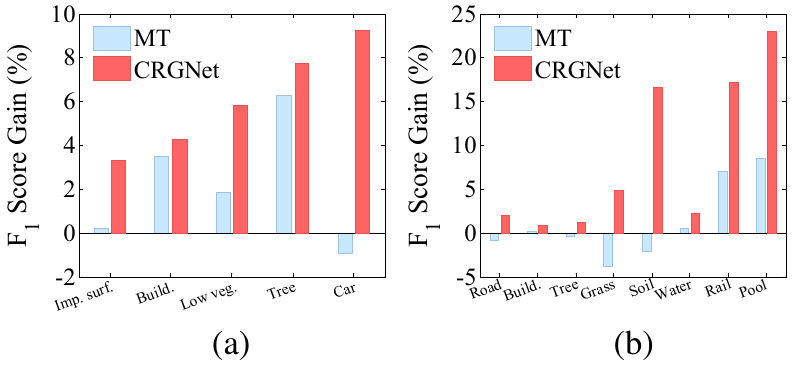}
  \caption{Per-class F$_1$ gains by the mean-teacher model (MT) and the proposed CRGNet. (a) Vaihingen. (b) Zurich Summer.}
\label{fig:Gain}
\end{figure}

We also visualize the prediction discrepancy between the base classifier $f_b$ and the expanded classifier $f_e$ in different iterations. As shown in \figurename~\ref{fig:dif}, there exist many red regions in the discrepancy maps for both datasets in the early period of the training (i.e., the $100$th iteration), which demonstrates that the predictions of the base classifier vary a lot from those of the expanded classifier. The reason for this phenomenon lies in the fact that the base classifier is trained with the original sparse point-level annotations, while the expanded classifier is supervised by the expanded annotations. However, as the iteration goes on, the discrepancy between predictions of both classifiers gets greatly reduced. It can be observed from the discrepancy map in the $1000$th iteration that most areas are colored blue in this case, which corresponds to low discrepancy values. These results also verify the effectiveness of the proposed consistency regularization.

We further make a detailed comparison of per-class F$_1$ gains between the mean-teacher (MT) model and the proposed CRGNet against the baseline method (vanilla DeepLab-v2). As shown in \figurename~\ref{fig:Gain}, there exist ``negative learning'' phenomenons on the car, road, grass, and soil categories in the MT method (blue bars). By contrast, the proposed CRGNet significantly mitigates this phenomenon (red bars). Besides, for those hard categories like the railway and the swimming pool in the Zurich Summer dataset, CRGNet can achieve remarkable F$_1$ gains over $15\%$. These results demonstrate the effectiveness of the proposed method in the weakly supervised semantic segmentation of VHR images.

\begin{table}[htb]
\caption{Mean F$_1$ scores with different values of $\tau$ (without self-training). Best Results Are Highlighted in \textbf{Bold}.}
\centering
\begin{tabular*}{\linewidth}{@{\extracolsep{\fill}}c|cccccc}
\toprule
$\tau$&0.3&0.5&0.8&0.9&0.95&0.99\\
\midrule
Vaihingen&64.11&64.79&64.81&64.73&\textbf{65.48}&65.21\\
Zurich Summer&67.21&67.33&68.42&68.37&\textbf{68.61}&68.15\\
\bottomrule
\end{tabular*}
\label{tab:tau}
\end{table}

\begin{table}[htb]
\caption{Mean F$_1$ scores with different values of $\lambda_{con}$ (without self-training). Best Results Are Highlighted in \textbf{Bold}.}
\centering
\begin{tabular*}{\linewidth}{@{\extracolsep{\fill}}c|cccccc}
\toprule
$\lambda_{con}$&0.03&0.1&0.3&1&3&10\\
\midrule
Vaihingen&64.12&64.77&65.31&65.48&\textbf{65.79}&65.13\\
Zurich Summer&67.30&67.81&67.57&\textbf{68.61}&68.14&66.05\\
\bottomrule
\end{tabular*}
\label{tab:lambda}
\end{table}

\subsection{Parameter Analysis}
In this subsection, we analyze how different values of the parameters in CRGNet would influence semantic segmentation performance.

\textit{The confidence threshold parameter $\tau$ in \eqref{eq:RG}}. Table \ref{tab:tau} shows that a smaller $\tau$ may not ensure the quality of the expanded annotations, while a larger $\tau$ may help to bring more accurate pseudo labels to the network. We empirically set $\tau$ as $0.95$ in the experiments for both datasets.

\textit{The consistency regularization weighting factor $\lambda_{con}$ in \eqref{eq:full}}. As shown in Table \ref{tab:lambda}, a too large $\lambda_{con}$ (i.e., $\lambda_{con}=10$) may bring about a too strong regularization for the training of both classifiers, which may be detrimental to the semantic segmentation performance in some cases. A good selection for $\lambda_{con}$ may range from $0.3$ to $3$. For both datasets, we empirically set $\lambda_{con}$ as $1$ in the experiments.

\subsection{Analysis of Different Loss Functions}
In the proposed CRGNet, we adopt the mean squared error (MSE) to implement the consistency regularization loss $\mathcal L_{con}$. Another intuitive choice is to use the KL divergence to constrain the probability distribution of $p_b$ and $p_e$. To explore how different loss functions would influence the performance of the proposed method, we further conduct experiments by changing the MSE loss to the KL divergence loss with temperature scaling \cite{he2020momentum}. The experimental results with different values of the temperature parameter $T$ are presented in Table \ref{tab:T_KL}. It can be observed from Table \ref{tab:T_KL} that a large value of the temperature $T$ would be harmful to the performance of the KL divergence. By contrast, a smaller $T$ would help to improve the performance. For example, on the Zurich Summer dataset, the mean $F_1$ score can reach $67.89\%$ when $T=0.01$, which is much higher than the one obtained with $T=1$ ($64.63\%$). Nevertheless, using the KL divergence loss with the temperature scaling technique does not show superior performance to the MSE loss used in the proposed method, as shown in Table \ref{tab:comparison}.

In fact, the phenomenon that the MSE loss may yield better results than the KL divergence loss for consistency regularization has been observed in previous research (although the KL divergence would seem a more natural choice). For example, in the classic mean-teacher method \cite{tarvainen2017mean}, Tarvainen \textit{et al.} conducted experiments and showed that MSE actually performs better than KL divergence (Fig. 4 (f) in \cite{tarvainen2017mean}). Readers can refer to Appendix C in \cite{tarvainen2017mean} for the theoretical analysis of how the MSE and the KL divergence would influence the performance of consistency regularization.

\begin{table}
\caption{Mean F$_1$ scores with different values of temperature parameter $T$ using the KL divergence loss (without self-training). Best Results Are Highlighted in \textbf{Bold}.}
\centering
\begin{tabular}{c|ccccc}
\toprule
$T$&0.01&0.1&1&10&100\\
\midrule
Vaihingen&64.33&\textbf{65.29}&64.91&63.18&61.93\\
Zurich Summer&\textbf{67.89}&66.58&64.63&62.80&63.88\\
\bottomrule
\end{tabular}
\label{tab:T_KL}
\end{table}

\begin{table}
\caption{Mean F$_1$ scores with different loss functions (without self-training). Best Results Are Highlighted in \textbf{Bold}.}
\centering
\begin{tabular}{c|cc}
\toprule
Loss Function&KL Divergence&Mean Squared Error\\
\midrule
Vaihingen&65.29&\textbf{65.48}\\
Zurich Summer&67.89&\textbf{68.61}\\
\bottomrule
\end{tabular}
\label{tab:comparison}
\end{table}

\section{Conclusions and Discussions}
This paper proposes a consistency-regularized region-growing network (CRGNet) for semantic segmentation of VHR remote sensing images using point-level annotations. The key idea of CRGNet is to iteratively select unlabeled pixels with high confidence to expand the annotated area from the original sparse points. However, directly learning from the expanded annotations may mislead the training of the network due to the potential errors in the region growing. To this end, we propose the consistency regularization strategy, where a base classifier and an expanded classifier are employed. Specifically, the base classifier is supervised by the original sparse annotations, while the expanded classifier aims to learn from the expanded annotations generated by the base classifier with the region-growing mechanism. The consistency regularization is thereby achieved by minimizing the discrepancy between the predictions from both the base and the expanded classifiers. We further conduct self-training with pseudo labels generated by the base classifier and the expanded classifier to finetune the proposed CRGNet. Extensive experiments on two challenging benchmark datasets demonstrate that the proposed CRGNet can yield competitive performance compared with the existing state-of-the-art approaches.

To analyze the contribution of each module in the proposed method, a detailed ablation study is further conducted. In both datasets, we find that directly applying the region growing mechanism only leads to limited mean F$_1$ gains, while combining it with the proposed consistency regularization can significantly improve the performance. The self-training technique and the post-processing of dCRF also play important roles in CRGNet. Although there still exists a performance gap between the weakly supervised methods and the fully supervised methods, our experiments demonstrate that the proposed CRGNet can help to close this performance gap while significantly reducing the annotation burden for dense pixel-wise labels. Thus, there may exist a great potential to apply the CRGNet to the practical scenarios where the dense pixel-wise annotations are difficult to collect.

Since the insufficiency of labeled data is a common challenge in many remote sensing tasks, whether the proposed consistency regularization strategy and the region-growing mechanism can yield good performance on other remote sensing scenarios is also worth studying. Besides, considering that the performance of the proposed CRGNet is largely determined by the quality of the expanded annotations, how to further improve the accuracy of the pseudo labels generated in the region-growing mechanism is a critical problem. We will try to explore these issues in our future work. While this study mainly focuses on the semantic segmentation with point-level annotations, other types of weak supervision like the scribble-level annotations and image-level annotations also deserve our attention. This could be another potential direction for researchers developing weakly supervised methods in the future.
\section*{Acknowledgment}
The authors would like to thank Prof. Vittorio Ferrari and Dr. Michele Volpi for providing the Zurich Summer dataset. The Vaihingen dataset was provided by the German Society for Photogrammetry, Remote Sensing and Geoinformation (DGPF).

\bibliographystyle{IEEEtran}

\bibliography{CRGNet}

\end{document}